\documentclass{article} 
\usepackage[preprint]{neurips_2020}
\usepackage{times}  
\usepackage{helvet} 
\usepackage{courier}  
\usepackage[hyphens]{url}  
\usepackage{graphicx} 
\usepackage{natbib}  
\usepackage{caption} 
\usepackage[utf8]{inputenc} 
\usepackage[T1]{fontenc}    
\usepackage{url}            
\usepackage{booktabs}       
\usepackage{amsfonts}       
\usepackage{microtype} 
\usepackage{graphicx}
\usepackage{amsmath}
\usepackage{amsthm}
\usepackage{booktabs}
\usepackage{algorithm}
\usepackage{algorithmic}
\usepackage{multirow}
\usepackage{subfigure}
\usepackage{diagbox}
\usepackage{amssymb}

\newtheorem{theorem}{Theorem}

\newtheorem{lemma}[theorem]{Lemma}
\usepackage[switch]{lineno}
\title{Memory and Computation-Efficient Kernel SVM via Binary Embedding and Ternary Model Coefficients}

\author{
  Zijian Lei\\
  Department of Computer Science, \\
  Hong Kong Baptist University, \\
  Hong Kong SAR, China\\
  cszjlei@comp.hkbu.edu.hk
  \And
  Liang Lan\\
  Department of Computer Science, \\
  Hong Kong Baptist University, \\
  Hong Kong SAR, China\\
  lanliang@comp.hkbu.edu.hk
}

\begin{document}
\maketitle
\begin{abstract}
Kernel approximation is widely used to scale up kernel SVM training and prediction. However, the memory and computation costs of kernel approximation models are still too high if we want to deploy them on memory-limited devices such as mobile phones, smartwatches, and IoT devices. To address this challenge, we propose a novel memory and computation-efficient kernel SVM model by using both binary embedding and binary model coefficients. First, we propose an efficient way to generate compact binary embedding of the data, preserving the kernel similarity. Second, we propose a simple but effective algorithm to learn a linear classification model with ternary coefficients that can support different types of loss function and regularizer. Our algorithm can achieve better generalization accuracy than existing works on learning binary coefficients since we allow coefficient to be $-1$, $0$, or $1$ during the training stage, and coefficient $0$ can be removed during model inference for binary classification. Moreover, we provide a detailed analysis of the convergence of our algorithm and the inference complexity of our model. The analysis shows that the convergence to a local optimum is guaranteed, and the inference complexity of our model is much lower than other competing methods. Our experimental results on five large real-world datasets have demonstrated that our proposed method can build accurate nonlinear SVM models with memory costs less than 30KB. 
\end{abstract}

\section{Introduction}
Kernel Support Vector Machine (SVM) is a powerful nonlinear classification model that has been successfully used in many real-world applications. Different from linear SVM, kernel SVM uses kernel function to capture the nonlinear concept. The prediction function of kernel SVM is $f(\mathbf{x}) = \sum_{\mathbf{x}_i}\alpha_ik(\mathbf{x}_i, \mathbf{x})$ 
where $\mathbf{x}_i$s are support vectors and $k(\mathbf{x}_i, \mathbf{x}_j)$ is a predefined kernel function to compute the kernel similarity between two data samples. Kernel SVM needs to explicitly maintain all support vectors for model inference. Therefore, the memory and computation costs of kernel SVM inference are usually huge, considering that the number of support vectors increases linearly with training data size on noisy data. To reduce the memory and computation costs of kernel SVM, many kernel approximation methods \cite{rahimi2008random, le2013fastfood, lan2019scaling, hsieh2014fast} have been proposed to scale up kernel SVM training and prediction in the past decade. The basic idea of these methods is to explicitly construct the nonlinear feature mapping $\mathbf{z} = \Phi(\mathbf{x}): \mathbb{R}^d \rightarrow \mathbb{R}^p$ such that $\mathbf{z}_i^\top\mathbf{z}_j \approx k(\mathbf{x}_i, \mathbf{x}_j)$ and then apply a linear SVM on $\mathbf{z}$. To obtain good classification accuracy, the dimensionality of nonlinear feature mapping $\mathbf{z}$ needs to be large. On the other hand, due to concerns on security, privacy, and latency caused by performing model inference remotely in the cloud, directly performing model inference on edge devices (e.g., Internet of Things (IoT) devices) has gained increasing research interests recently \cite{kumar2017resource, kusupati2018fastgrnn}. Even though those kernel approximation methods can significantly reduce the memory and computation costs of exact kernel SVM, their memory and computation costs are still too large for on-device deployment. For example, the Random Fourier Features (RFE), which is a very popular kernel approximation method for large scale data, requires $\sim d\times p \times 32$ bits for nonlinear feature transformation, $p\times 32$ bits for nonlinear feature representation, and $p\times c \times 32$ bits ($c$ is the number of classes) for classification model to predict the label of a single input data sample. The memory cost of RFE can easily exceed several hundred megabytes (MB), which could be prohibitive for on-device model deployment.

Recently, binary embedding methods \cite{yu2017binary, needell2018simple} have been widely used for reducing memory cost for data retrieval and classification tasks. The nice property of binary embedding is that each feature value can be efficiently stored using a single bit. Therefore, binary embedding can reduce the memory cost by 32 times compared with storing full precision embedding. The common way for binary embedding is to apply $\text{sign}(\cdot)$ function to the famous Johnson-Lindenstrauss embeddings: random projection embedding justified by the Johnson-Lindenstrauss Lemma \cite{johnson1986extensions}. It obtains the binary embedding by $\mathbf{z} = \text{sign}(\mathbf{R}\mathbf{x})$ where $\mathbf{R} \in \mathbb{R}^{d\times p}$ is a random Gaussian matrix \cite{needell2018simple,ravi2019efficient} or its variants \cite{yu2017binary, gong2012angular, shen2017classification} and $\text{sign}(\cdot)$ is the element-wise sign function. However, most of them focus on data retrieval, and our interest is on-device nonlinear classification. The memory cost of the full-precision dense matrix $\mathbf{R}$ is large for on-device model deployment. Besides, even though $\text{sign}(\mathbf{R}\mathbf{x})$ provide a nonlinear mapping because of the $\text{sign}(\cdot)$ function, the theoretical analysis on binary mapping $\text{sign}(\mathbf{Rx})$ can only guarantee that the angular distance among data samples in the original space is well preserved. It cannot well preserve the kernel similarity among data samples, which is crucial for kernel SVM. 

In this paper, we first propose a novel binary embedding method that can preserve the kernel similarity of the shift-invariant kernel. Our proposed method is built on Binary Codes for Shift-invariant kernels (BCSIK) \cite{raginsky2009locality} but can significantly reduce the memory and computation costs of BCSIK. Our new method can reduce the memory cost of BCSIK from $O(dp)$ to $O(p)$ and the computational cost of BCSIK from $O(dp)$ to $O(plogd)$. Second, in addition to binary embedding, we propose to learn the classification model with ternary coefficients. The classification decision function is reduced to a dot product between a binary vector and a ternary vector, which can be efficiently computed. Also, the memory cost of the classification model will be reduced by 32 times. Unlike existing works \cite{shen2017classification, AlizadehFLG19an} on learning binary coefficients, we allow the model coefficient to be $\{-1, 0, 1\}$ during the training stage. This additional 0 can help to remove uncorrelated binary features and improve the generalization ability of our model. The $0$ coefficients and corresponding transformation column vectors in matrix $\mathbf{R}$ can be safely removed during the model inference for binary classification problems. A simple but effective learning algorithm that can support different types of the loss function and regularizer is proposed to learn binary model coefficients from data. Third, we provide a detailed analysis of our algorithm's convergence and our model's inference complexity. The analysis shows that the convergence to a local optimum is guaranteed, and the memory and computation costs of our model inference are much lower than other competing methods. We compare our proposed method with other methods on five real-world datasets. The experimental results show that our proposed method can greatly reduce the memory cost of RFE while achieves good accuracy. 

\section{Methodology\label{Methodology}}
\subsection{Binary Embedding for Shift-Invariant Kernels}
\textbf{Preliminaries on Binary Codes for Shift-Invariant Kernels (BCSIK).} BCSIK \cite{raginsky2009locality} is a random projection based binary embedding method which can well preserve the kernel similarity defined by a shift-invariant kernel (e.g., Gaussian Kernel). It works by composing random Fourier features with a random sign function. Let us use $\mathbf{x} \in \mathbb{R}^{d}$ to denote an input data sample with $d$ features. BCSIK encodes $\mathbf{x}$ into a $p$-dimensional binary representation $\mathbf{z}$ as
\begin{equation}\label{BCSIK}
\mathbf{z} = \text{sign}(\text{cos}(\mathbf{R}^{\top}\mathbf{x} + \mathbf{b}) + \mathbf{t}),
\end{equation}
where each column $\mathbf{r}$ in matrix $\mathbf{R}\in \mathbb{R}^{d \times p}$ is randomly drawn from a distribution corresponding to an underlying shift-invariant kernel. For example, for Gaussian kernel $k(\mathbf{x}_i, \mathbf{x}_j) = \text{exp}(-\frac{\|\mathbf{x}_i-\mathbf{x}_j\|}{2\sigma^2})$, each entry in $\mathbf{r}$ is drawn from a normal distribution $\mathcal{N}(0, \sigma^{-2})$. $\mathbf{b} \in \mathbb{R}^{p}$ is a column vector where each entry is drawn from a uniform distribution from $[0, 2\pi]$. $\mathbf{t} \in \mathbb{R}^{p}$ is a column vector where each entry is drawn from a uniform distribution from $[-1, 1]$. cos($\cdot$) and sign($\cdot$) are the element-wise cosine and sign functions. As can be seen, $\Phi(\mathbf{x}) = \text{cos}(\mathbf{R^\top}\mathbf{x} + \mathbf{b})$ in (\ref{BCSIK}) is the random Fourier features for approximating kernel mapping which has the theoretical guarantee $\mathbb{E}[\Phi(\mathbf{x}_1)^\top\Phi(\mathbf{x}_2)] = k(\mathbf{x}_1, \mathbf{x}_2)$ for shift-invariant kernel \cite{rahimi2008random}, where $\mathbb{E}$ is the statistical expectation. $\text{sign}(\Phi(\mathbf{x}) + \mathbf{t})$ uses random sign function to convert the full-precision mapping $\Phi(\mathbf{x})$ into binary mapping $\mathbf{z}$. Each entry in $\mathbf{z}$ can be stored efficiently using a single bit. Therefore, compared with RFE, BCSIK can reduce the memory cost of storing approximated kernel mapping by 32 times.       

Besides memory saving, a great property of BCSIK is that the normalized hamming distance between the binary embedding of any two data samples sharply concentrates around a well-defined continuous function of the kernel similarity between these two data samples shown in the Lemma \ref{lemma:BCSIK}. 

\begin{lemma} [Johnson-Lindenstrauss Type Result on BCSIK \cite{raginsky2009locality}]\label{lemma:BCSIK}
Define the functions $h_1(u) \triangleq \frac{4}{\pi^2}(1-u)$ and $h_2(u) \triangleq min\{\frac{1}{2}\sqrt{1-u}, \frac{4}{\pi^2}(1-\frac{2}{3}u)\}$, where $u \in [0, 1]$. Fix $\epsilon, \delta \in (0, 1)$. Then for any finite dataset $\{\mathbf{x}_1, \dots, \mathbf{x}_n\}$ of $n$ data samples in $\mathbb{R}^d$, the following inequality about the normalized hamming distance on the binary embedding 
between any two data samples $\mathbf{z}_i$ and $\mathbf{z}_j$ holds true with probability $\geq 1 - \epsilon$ with $p \geq \frac{1}{2\delta^2}log(\frac{n^2}{\epsilon})$
\begin{equation}\label{Lemma_JL_type}
h_1(k(\mathbf{x}_i, \mathbf{x}_j)) - \delta \leq \frac{1}{p}d_{H}(\mathbf{z}_i, \mathbf{z}_j) \leq h_2(k(\mathbf{x}_i, \mathbf{x}_j) + \delta.
\end{equation}
\end{lemma}
Note that the hamming distance between $\mathbf{z}_i$ and $\mathbf{z}_j$ can be expressed as $d_{H}(\mathbf{z}_i, \mathbf{z}_j) = \frac{1}{2}(p - \mathbf{z}_i^\top\mathbf{z}_j)$. The bounds in Lemma \ref{lemma:BCSIK} indicate that the binary embedding $\mathbf{z} \in \{-1,1\}^{p}$ as defined in (\ref{BCSIK}) well preserves the kernel similarity obtained from the underlying shift-invariant kernel.

\textbf{Reduce the Memory and Computation Costs of BCSIK.} When considering on-device model deployment, the bottleneck of obtaining binary embedding is the matrix-vector multiplication $\mathbf{R}^{\top}\mathbf{x}$ as shown in (\ref{BCSIK}). It requires $O(dp)$ time and space. By considering $p$ is usually several times larger than $d$ for accurate nonlinear classification, the memory cost of storing $\mathbf{R}$ could be prohibitive for on-device model deployment. Therefore, we propose to generate the Gaussian random matrix $\mathbf{R}$ using the idea of Fastfood \cite{le2013fastfood}. The core idea of Fastfood is to reparameterize a Gaussian random matrix by a product of Hadamard matrices and diagonal matrices. Assuming $d=2^q$\footnote{We can ensure this by padding zeros to original data} and $q$ is any positive integer, it constructs $\mathbf{V} \in \mathbb{R}^{d \times d}$ as follows to reparameterize a $d \times d$ Gaussian random matrix,
\begin{equation}\label{eq:V}
    \mathbf{V} = \frac{1}{\sigma\sqrt{d}}\mathbf{SHG\Pi HB},
\end{equation}
where
\begin{itemize}
    \item $\mathbf{S,G}$ and $\mathbf{B}$ are diagonal matrices. $\mathbf{S}$ is a random scaling matrix, $\mathbf{G}$ has random Gaussian entries and $\mathbf{B}$ has elements are independent random signs $\{-1,1\}$. The memory costs of $\mathbf{S,G}$ and $\mathbf{B}$ are $O(d)$.
    \item $\mathbf{\Pi} \in \{0,1\}^{d\times d}$ is a random permutation matrix and also has $O(d)$ space complexity
    \item $\mathbf{H} \in \mathbb{R}^{d \times d}$ is the Walsh-Hadamard matrix defined recursively as:  \\
	$   \mathbf{H}_d = \left[\begin{array}{cc}
	\mathbf{H}_{d/2} & \mathbf{H}_{d/2} \\
	\mathbf{H}_{d/2} & -\mathbf{H}_{d/2}
	\end{array}
	\right]$
	with
	$     \mathbf{H}_2 = \left[\begin{array}{cc}
	1 & 1 \\
	1 & -1
	\end{array}
	\right]$; The fast Hadamard transform allows us to compute $\mathbf{Hx}$ in $O(d\log d )$ time  using the Fast Fourier Transform (FFT) operation.
\end{itemize}

When reparameterizing a $d \times p$ Gaussian random matrix ($p \gg d$), Fastfood replicates (\ref{eq:V}) for $p/d$ independent
random matrices $\mathbf{V}_i$ and stack together as
\begin{equation}\label{eq:tilde_W}
    \mathbf{\Tilde{R}}^\top =[\mathbf{V}_1; \mathbf{V}_2; \dots, \mathbf{V}_{p/d}]^\top,
\end{equation}
 until it has enough dimensions. Then, we can generate the binary embedding in a memory and computation-efficient way as follows,
 \begin{equation}\label{BCSIK_fastfood}
\mathbf{z} = \text{sign}(\text{cos}(\mathbf{\Tilde{R}}^{\top}\mathbf{x} + \mathbf{b}) + \mathbf{t}),
\end{equation}
Note that each column in $\mathbf{\Tilde{R}}$ is a random Gaussian vector from $\mathcal{N}(0, \sigma^{-2}\mathbf{I}_d)$ as proved in \cite{le2013fastfood}, therefore the Lemma \ref{lemma:BCSIK} still holds when we replace $\mathbf{R}$ in (\ref{BCSIK}) by $\mathbf{\Tilde{R}}$ for efficiently generating binary embedding. Note that the Hadamard matrix does not need to be explicitly stored. Therefore, we can reduce the space complexity of (\ref{BCSIK}) in BCSIK from $O(dp)$ to $O(p)$ and reduce the time complexity of (\ref{BCSIK}) from $O(dp)$ to $O(plog(d))$. 

\subsection{Ternary Model Coefficients for Classification}
By using (\ref{BCSIK_fastfood}), each data sample $\mathbf{x}_i \in \mathbb{R}^d $ is transformed to a bit vector $\textbf{z}_i \in \mathbb{R}^p$. Then, we can train a linear classifier on $\{\mathbf{z}_i, y_i\}_{i=1}^{n}$ to approximate the kernel SVM. Suppose the one-vs-all strategy is used for multi-class classification problems, the memory cost of the learned classifier is $p \times c \times 32$, where $c$ is the number of classes. Since $p$ usually needs to be very large for accurate classification, the memory cost of the classification model could also be too huge for edge devices, especially when we are dealing with multi-class classification problems with a large number of classes. 

In here, we propose to learn a classification model with ternary coefficient for reducing the memory cost of classification model. Moreover, by using binary model coefficients, the decision function of classification is reduced to a dot product between two binary vectors which can be very efficiently computed. Compare to existing works \cite{shen2017classification, AlizadehFLG19an} on learning binary model coefficients which constrain the coefficient to be 1 or $-1$, we allow the model coefficients to be 1, 0 or $-1$ during the training stage. The intuition of this operation came from two aspect. First, suppose our data distribute in a hyper-cube and linearly separable, the direction of the classification hyper-plane can be more accurate when we allow ternary coefficients. For example, in lower projected dimension, the binary coefficients can achieve $2^p$ direction while the ternary ones can achieve $3^p$.  This additional value $0$ can help to remove uncorrelated binary features and improve the generalization ability of our model as the result of importing the regularization term. In addition, we add a scaling parameter $\alpha$ to prevent possible large deviation between full-precision model coefficients and quantized model coefficients which affects the computation of loss function on training data. Therefore, our objective is formulated as
\begin{equation}\label{eq:ternerizedModel}
	\begin{split}
	\min\limits_{\alpha, \mathbf{w}}& \frac{1}{n}\sum_{i=1}^{n}\ell(y_i, \alpha\mathbf{w}^{\top}\mathbf{z}_i)) + \lambda R(\mathbf{w})\\
	\text{s.t.} & \ \ \ \ \mathbf{w} \in \{-1, 0, 1\}^{p}\\
	& \ \ \ \ \alpha > 0.
	\end{split}
\end{equation}
where $p$ is dimensionality of $\textbf{z}_i$ and $y_i$ is the corresponding label for $\textbf{z}_i$. $\ell(y_i, \alpha\mathbf{w}^{\top}\mathbf{z}_i)$ in (\ref{eq:ternerizedModel}) denotes a convex loss function and $R(\mathbf{w})$ denotes a regularization term on model parameter $\mathbf{w}$. $\lambda$ is a hyperparameter to control the tradeoff between training loss and regularization term.

\subsubsection{Learning ternary coefficient with Hinge Loss and $l_2$-norm Regularizer}
For simplicity of presentation, let us assume $y_i \in \{-1, 1\}$. Our model can be easily extended to multi-class classification using the one-vs-all strategy. Without loss of generality, in this section, we show how to solve (\ref{eq:ternerizedModel}) when hinge loss and $l_2$-norm regularizer are used. In other word, $\ell(y_i, \alpha\mathbf{w}^{\top}\mathbf{z}_i)$ is defined as $\max(0, 1 - y_i\alpha\mathbf{w}^{\top}\mathbf{z}_i)$. $R(\mathbf{w})$ is defined as $\alpha^2\sum_{j=1}^{p}w_j^2$. Any other loss function or regularization can also be applied.

By using hinge loss and $l_2$-norm regularization, then (\ref{eq:ternerizedModel}) will be rewritten as:
\begin{equation}\label{eq:bm_hingloss}
	\begin{split}
	\min\limits_{\alpha, \mathbf{w}}& \frac{1}{n}\sum_{i=1}^{n}\max(0, 1 - \alpha y_i\mathbf{w}^{\top}\mathbf{z}_i) + \lambda\alpha^2 \sum_{j=1}^{p}w_j^2\\
	\text{s.t.} & \ \ \ \ \mathbf{w} \in \{-1, 0, 1\}^{p}\\
	& \ \ \ \ \alpha > 0.
	\end{split}
\end{equation}

We can use alternating optimization to solve (\ref{eq:bm_hingloss}): (1) fixing $\mathbf{w}$ and solving $\alpha$; and (2) fixing $\alpha$ and solving $\mathbf{w}$.

\noindent \textbf{1. Fixing $\mathbf{w}$ and solving $\alpha$}. When $\mathbf{w}$ is fixed, (\ref{eq:bm_hingloss}) will be reduced to a problem with only one single variable $\alpha$ as follows
\begin{equation}\label{eq:bm_hingloss_alpha}
	\begin{split}
	\min\limits_{\alpha} & \frac{1}{n}\sum_{i=1}^{n}\max(0, 1 -  (y_i\mathbf{w}^{\top}\mathbf{z}_i)\alpha) + (\lambda \sum_{j=1}^{p}w_j^2)\alpha^2\\
	\text{s.t.} & \ \ \ \ \alpha > 0.
	\end{split}
\end{equation}
It is a convex optimization problem with non negative constraint \cite{boyd2004convex}.

\noindent \textbf{2. Fixing $\alpha$ and solving $\mathbf{w}$}. When $\alpha$ is fixed, (\ref{eq:bm_hingloss}) will change to the following optimization problem, 
\begin{equation}\label{eq:bm_hingloss_w}
	\begin{split}
	\min\limits_{\mathbf{w}}& \frac{1}{n}\sum_{i=1}^{n}\max(0, 1 - \alpha y_i\mathbf{w}^{\top}\mathbf{z}_i) + \lambda\alpha^2 \sum_{j=1}^{p}w_j^2\\
	\text{s.t.} & \ \ \ \ \mathbf{w} \in \{-1, 0, 1\}^{p}.\\
	\end{split}
\end{equation}

Due to the non-smooth constraints, minimizing (\ref{eq:bm_hingloss}) is an NP-hard problem and needs $O(3^{p})$ time to obtain the global optimal solution. The Straight Through Estimator (STE) \cite{bengio2013estimating} framework which is popular for learning binary deep neural networks can be used to solve (\ref{eq:bm_hingloss}). However, by considering that the STE can be unstable near certain local minimum \cite{yin2019understanding, liu2019learning}, we propose a much simpler but effective algorithm to solve (\ref{eq:bm_hingloss}). The idea is to update parameter $\mathbf{w}$ bit by bit, i.e., one bit each time for $w_j, j = 1, \dots, p$, while keep other $(p-1)$ bits fixed. Let use $\textbf{w}_{(\neg j)}$ to denote the vector that equals to \textbf{w} except the $j$-th entry is set to 0.  Therefore, (\ref{eq:bm_hingloss}) will be decomposed to a series of subproblems. A subproblem of (\ref{eq:bm_hingloss}) which only involves a single variable $w_j$ can be written as (\ref{eq:bm_hingeloss_w_j}) and $\alpha = \alpha_w$.
\begin{equation}\label{eq:bm_hingeloss_w_j}
	\begin{split}
	\min\limits_{w_j} \ L =& \frac{1}{n}\sum_{i=1}^{n}\max(0, 1 - y_i\alpha(\mathbf{w}_{\neg j}^\top\mathbf{z}_i + w_jz_{ij}))\\
	&+ \lambda\alpha^2 w_j^2 + \lambda \alpha^2\mathbf{w}_{\neg j}^\top\mathbf{w}_{\neg j} \\
	\text{s.t.} & \ \ \ \ w_j \in \{-1, 0, 1\}. \\
	\end{split}
\end{equation}

Since $w_j \in \{-1, 0, 1\}$, objective (\ref{eq:bm_hingeloss_w_j}) can be solved by just enumerating all three possible values $\{-1, 0, 1\}$ for $w_j$ and select the one with the minimal objective value. Note that $\mathbf{w}^\top\mathbf{z}_i$ and $\mathbf{w}^\top\mathbf{w}$ can be pre-computed. Then, in each subproblem (\ref{eq:bm_hingeloss_w_j}), both $\mathbf{w}_{\neg j}^\top\mathbf{z}_i = \mathbf{w}^\top\mathbf{z}_i - w_jz_{ij}$ and $\mathbf{w}_{\neg j}^\top\mathbf{w}_{\neg j} = \mathbf{w}^\top\mathbf{w} - w_j^2$ can be computed in $O(1)$ time. Therefore, we only need $O(n)$ time to evaluate the $L(w_j = -1)$, $L(w_j = 0)$ and $L(w_j = 1)$ for (\ref{eq:bm_hingeloss_w_j}). The optimal solution for each subproblem,
\begin{equation}\label{eq:solution_w_j}
w_j^{*} = \text{argmin}\{L(w_j = -1), L(w_j = 0), L(w_j = 1)\}
\end{equation}

can be obtained in $O(n)$ time. Then $w_j$ will be updated to $w_j^{*}$ if it is not equal to $w_j^{*}$. This simple algorithm can be easily implemented and applied to other popular loss functions (e.g., logloss, square loss) and regularizers (e.g., $l_1$ regularizer). Note that for some specific loss function (e.g., square loss), a close form solution for $w_j$ can be derived without using enumeration as shown in (\ref{eq:solution_w_j}).

\textbf{Parameter Initialization.} In this section, we propose a heuristic way to initialize $\alpha$ and $\mathbf{w}$ which can help us to quickly obtain a local optimal solution for (\ref{eq:bm_hingloss}). The idea is we first randomly select a small subset of training data to apply the linear SVM on transformed data $\{\mathbf{z}_i, y_i\}_{i=1}^{m}$ to get the full-precision solution $\mathbf{w}_{full}$. Then the parameter $\mathbf{w}$ is initialized as
\begin{equation}\label{eq:initialization}
 \begin{split}
  \mathbf{w} &= \text{sign}(\mathbf{w}_{full}) \\
  \alpha &= \frac{\|\mathbf{w}_{full}\|_1}{p}.\\
 \end{split}
 \end{equation}
Empirically this initialization can lead to a fast convergence to a local optimum and produce better classification accuracy. The results in show in Figure \ref{fig:intialization} and will be discussed in detail in the experiments section. 
\subsubsection{Efficiently Compute Ternary-Binary Dot Product.}
Once we get the ternary coefficients and the binary feature embedding, the following question is how to efficiently compute $\mathbf{w}^T\mathbf{z}$ since the parameter $\alpha$ only scale the value $\mathbf{w}^T\mathbf{z}$ and will not affect the prediction results. Next, we will discuss efficiently computing $\mathbf{w}^T\mathbf{z}$ using bit-wise operations. 
 
For binary classification problem, after training stage, we can remove the coefficient $w_j$s with zero value and also the corresponding columns in matrix $\mathbf{B}$. Therefore, for model deployment, our classification model $\mathbf{w}$ is still a binary vector and only one bit is needed for storing one entry $w_j$. This is different to ternary weight networks \cite{li2016ternary} where the coefficient $0$ needs to be explicitly represented for model inference. The predict score can be computed as, 
  \begin{equation}
     \mathbf{w}^T\mathbf{z} = 2\text{POPCOUNT}(\mathbf{z}\;\text{XNOR}\;\mathbf{w}) - \text{len}(\mathbf{z}),
 \end{equation}
 where \text{XNOR} is a bit-wise operation, $\text{POPCOUNT}(\mathbf{a})$ returns the number of 1 bits in $\mathbf{a}$ and the $\text{len}(\mathbf{z})$ returns the length of vector $\mathbf{z}$.
  
 For multi-class classification problem, the columns in $\mathbf{B}$ can be only removed if their corresponding coefficients are zero simultaneously in all $\mathbf{w}$ vectors for all $c$ classes. Therefore we will need 2-bits to store the remaining coefficients after dropping the 0 value simultaneously in all classes. However, this 2-bit representation might have little influence in computational efficiency. For a given class $j$ with model parameter $\mathbf{w}$,  the original $\mathbf{w}_j$ with ternary values can be decompose as $\mathbf{w} = \mathbf{w}^{p} \odot \mathbf{w}^{s}$ where $\mathbf{w}^{s}$ and $\mathbf{w}^{p}$ are defined as 
 \begin{equation}\label{coefficeints_decompose}
 {w}^{p}_j =  \left\{\begin{array}{l}
         1\qquad \text{if}\ {w}_j = 1 \\
         -1 \quad \text{otherwise}
    \end{array},   \right. \;
     {w}^{s}_j =  \left\{\begin{array}{l}
     1 \quad \text{if}\ {w}_j = \pm 1\\
         0 \quad {w}_j = 0 
    \end{array}   \right.
 \end{equation}
 Then $\mathbf{w}^T\mathbf{z}$ can be compute as:
 \begin{equation}
     \mathbf{w}^T\mathbf{z} = 2\text{POPCOUNT}(\mathbf{z}\;\text{XNOR}\;\mathbf{w}^{p}\;\text{AND}\;\mathbf{w}^{s}) - \text{len}(\mathbf{z}).
 \end{equation}
Note that for 1 in (\ref{coefficeints_decompose}) represent logic TRUE and store as 1, while 0 and $-1$ in (\ref{coefficeints_decompose})  will  be stored as 0 to represent logic FALSE in model inference.

\begin{algorithm}[!h]
\caption{Memory and computation-efficient kernel SVM via binary embedding and ternary model coefficients}
\begin{algorithmic}
	\STATE \underline{\textbf{Training}}
	\STATE \textbf{Input}:	training data set $D = \{\mathbf{x}_i, y_i\}_{i=1}^{n}$, new dimension $p$, regularization parameter $\lambda$;
    \STATE \textbf{Output}: transformation parameter $\mathbf{\Tilde{R}} \in \mathbb{R}^{d\times p}$, $\mathbf{b} \in \mathbb{R}^{p}$, $\mathbf{t} \in \mathbb{R}^{p}$; classification model with ternary coefficients $\mathbf{w}$;
	\end{algorithmic}
	\begin{algorithmic}[1]
		\STATE Generate random Gaussian matrix $\mathbf{\Tilde{R}}$ as defined in (\ref{eq:tilde_W}), random vector $\mathbf{b}$ and $\mathbf{t}$.
		\STATE Compute the binary embedding as defined in (\ref{BCSIK_fastfood}) 
		\STATE Initialization: $\mathbf{w}$, $\alpha$ as shown in (\ref{eq:initialization})
        \STATE $r  =\mathbf{w}^\top\mathbf{w}$. \ \ \ \ \ \ \ \ \ \ \ \ \ \ \ \ \ \ \ \# Pre-Computing regularizer 
		\FOR{\text{$i = 1$ to $n$}}  
             \STATE $h_i = \mathbf{w}^\top\mathbf{z}_i$ \ \ \ \ \ \ \ \ \ \ \ \ \ \ \ \# Pre-Computing predictions
        \ENDFOR
        \REPEAT
	
        \STATE solve (\ref{eq:bm_hingloss_alpha}) by fixing $\mathbf{w}$ 
        \REPEAT [Learning $\mathbf{w}$ by fixing $\alpha$]
              \FOR{\text{$j = 1$ to $p$}}
                  \STATE evaluate $L(w_j=-1)$, $L(w_j=0)$ and $L(w_j=1)$
                  \STATE obtain the optimal solution of $w_j^{*}$ based on (\ref{eq:solution_w_j})
                  \IF{$w_j \neq w_j^{*}$}
                      \STATE set $w_j^{old} = w_j$ and update $w_j = w_j^{*}$
                      \STATE update $h_i = h_i - (w_j^{old} - w_j)z_{ij}$ for each  sample
                  \STATE update $r = r - (w_j^{old})^2 + (w_j)^2$
                  \ENDIF
              \ENDFOR
       \UNTIL{objective (\ref{eq:bm_hingloss_w}) converge;}
       \UNTIL{objective (\ref{eq:bm_hingloss}) converge;}
	\end{algorithmic}
	\begin{algorithmic}
		\STATE \underline{\bf{Prediction}} \STATE
		\textbf{Input}: a test sample $\mathbf{x}_i$, ternary coefficient $\mathbf{w}$, transformation parameters $\mathbf{\Tilde{R}}, \mathbf{b}, \mathbf{t} $;
		\STATE \textbf{Output}: predicted label $\hat{y}_i$;
		
	\end{algorithmic}
	\begin{algorithmic}[1]
 		\STATE compute binary embedding $\mathbf{z}_i$ as defined in (\ref{BCSIK_fastfood}) 
 		\STATE obtain the predicted label by 
  		$\hat{y}_i = \text{sign}(\mathbf{w}^\top\mathbf{z}_i)$;
	\end{algorithmic}
    \label{alg:bKSVM}
\end{algorithm}
  
\subsection{Algorithm Implementation and Analysis}
We summarize our proposed algorithm for memory and computation-efficient kernel SVM in Algorithm \ref{alg:bKSVM}. During the training stage, step 2 needs $O(np)$ space and $O(nplog(d))$ time to obtain the binary embedding by using FFT. To learn the binary model coefficients (from step 4 to step 21), it requires $O(tnp)$ time where $t$ is the number of iterations. Therefore, the training process can be done efficiently. Our algorithm can be easily implemented and applicable to other loss functions and regularizers. The source code of our implementation is included in the supplementary materials and will be publicly available. Next, we present a detailed analysis of the convergence of our algorithm and the inference complexity of our model. 
\begin{lemma}[Convergence of Algorithm 1]\label{Lemma:convergence} The Algorithm 1 will converge to a local optimum of objective (\ref {eq:bm_hingloss}). 
\end{lemma}
The proof of of Lemma \ref{Lemma:convergence} can be done based on the fact that each updating step in solving (\ref{eq:bm_hingloss_alpha}) and (\ref{eq:solution_w_j}) will only decrease the objective function (\ref{eq:bm_hingloss}). By using our proposed parameter initialization method, Our experimental results empirically show that our proposed algorithm can converge to a local optimal solution fast.  

\begin{table}[t]
		\centering
		\caption{Memory and Computation Costs for Different Algorithms}
		\begin{tabular}{|c|c|c|c|c|l|}
		\hline
        \multirow{2}{*}{Algorithms} & \multicolumn{3}{c|}{Memory Cost (bits)} & \multicolumn{2}{c|}{Computation Cost} \\
        \cline{2-6}
         & Transformation & Embedding & Classifier & \# of BOPS & \# of FLOPs\\
        \hline
        \hline
	    RFE  & $\sim d \times p \times 32$  & $p\times32$ & $c\times p\times32$ & $-$ & $O(d \times p + p)$  \\
        \hline
        Fastfood & $\sim p\times 5 \times 32 $ & $p\times32$ & $c\times p\times32$ & $-$ & $O(plog(d) + p)$\\
        \hline
        BJLE  & $\sim d \times p \times 32$   & $p$  & $c\times p\times32$ & $-$ & $O(d \times p + p)$  \\
        \hline
        BCSIK  & $\sim d \times p \times 32$   & $p$  & $c\times p\times32$ & $-$ & $O(d \times p + p)$  \\
        \hline
        Our Proposed  &  $\sim p\times 5 \times 32$ & $p$ & $c\times p$ & $O(p)$ &  $O(plog(d))$  \\
		\hline
	\end{tabular}
    \label{cost_different_algorithms}
\end{table}

\textbf{Inference Complexity of Our Model} The main advantage of our model is that it provides memory and computation-efficient model inference. To compare the memory and computation costs for classifying a single input data sample with other methods, we decompose the memory cost into (1) memory cost of transformation parameters; (2) memory cost of embedding; and (3) memory cost of the classification model. We also decompose the computation cost into (1) number of binary operations (\# of BOPS) and (2) number of float-point operations (\# of FLOPS). To deploy our model, since we do not need to store the Hadamard matrix $\mathbf{H}$ explicitly, we only need $32 \times 3 \times p$ bits to store $\mathbf{S}, \mathbf{G}, \mathbf{\Pi}$. $\mathbf{B}$ can be stored in $p$ bits, $\mathbf{b}$ and $\mathbf{t}$ can be stored in $2\times 32 \times p$ bits. Therefore, total $\sim p\times 5 \times 32 $ bits are need for storing transformation parameters. For RFE, BCSIK, and Binary Johnson-Lindenstrauss Embedding (BJLE), they need to maintain a large $d\times p$ Gaussian matrix explicitly. Therefore, their memory cost of transformation parameters is $\sim d\times p \times 32$ bits, which is $\sim d$ times larger than our proposed method. As for storing the transformed embedding, RFE needs $p\times 32$ bits where binary embedding methods (i.e., BJLE, BCSIK, and our proposed method) only need $p$ bits. As for storing the classification model, assume that the number of classes is $c$, and one-vs-all strategy is used. Then, $c \times p \times 32$ bits are needed for full-precision model coefficients, and $c\times p$ bits are needed for binary model coefficients. With respect to the computation complexity of our model, step 1 in prediction needs $O(dlog(d))$ FLOPs, and step 2 needs $O(p)$ BLOPs. The memory and computation costs for different algorithms are summarized in Table \ref{cost_different_algorithms}.
 Furthermore, since both $\mathbf{w}$ and $\mathbf{z}_i$ are bit vectors, the dot product between them can be replaced by cheap XNOR and POPCOUNT operations, which have been showing to provide $58\times$ speed-ups compared with floating-point dot product in practice \cite{rastegari2016xnor}.

\begin{table*}[htb]
\centering
\caption{Accuracy and memory cost of different models}
\begin{tabular}{c|l|c|c|c|c|c}
\hline
& metric &{\sf usps} & {\sf covtype} &{\sf webspam }&{\sf mnist} & {\sf fashion-mnist} \\
\hline
\multirow{3}{*}{RFE} & accuracy &98.63 &84.19 &97.79  & 96.88 &87.5 \\
& memory cost &2136KB & 448KB & 2048KB & 6328KB & 6328KB  \\
& memory red. &1x     & 1x    & 1x     & 1x     & 1x
\\
\hline
\multirow{3}{*}{ Fastfood}  & accuracy & 98.29 &84.82&97.47 &96.48 & 87.55\\
& memory cost &112KB  & 40KB & 40KB & 112KB & 112KB \\
& memory red. &19x     & 11x    & 51x     & 57x     & 57x \\
\hline
\hline
\multirow{3}{*}{BJLE}  & accuracy & 97.53  &80.77&96.39 & 92.95&84.06 \\
& memory cost &2128KB& 440KB & 2040KB & 6320KB & 6320KB \\
& memory red. &1x     & 1x    & 1x     & 1x     & 1x \\
\hline
\multirow{3}{*}{ BCSIK}  & accuracy &97.85   &81.91 &96.41 & 93.05&84.01 \\
& memory cost &2128KB & 440KB & 2040KB & 6320KB & 6320KB \\
& memory red. &1x     & 1x    & 1x     & 1x     & 1x \\
\hline
\hline
\multirow{3}{*}{ Our Method}  & accuracy &98.01 & 81.68& 96.34& 93.27& 83.29\\ 
& memory cost  &104KB & 32KB & 32KB & 104KB & 104KB \\
& memory red.       &21x     & 14x    & 64x     & 61x     & 61x \\
\hline
\multirow{3}{*}{Our Method-b }  & accuracy & 96.57 &78.12 &94.75 &92.66 & 82.07\\
& memory cost &29KB & 24KB & 24KB & 29KB & 29KB \\
& memory red. &74x     & 19x    & 85x     & 218x     & 218x \\
\hline
    \end{tabular}

    \label{tab:results}
\end{table*}

\section{Experiments\label{experiments}}
In this section, we compare our proposed method with other efficient kernel SVM approximation methods and binary embedding methods. We evaluate the performance of the following six methods.
\begin{itemize}
\setlength{\topsep}{0pt}
\setlength{\itemsep}{0.5pt}
\setlength{\parsep}{0pt}
\setlength{\parskip}{0pt}
\item Random Fourier Features (RFE) \cite{rahimi2008random}: It approximates the shift-invariant kernel based on its Fourier transform.
\item Fastfood kernel  (Fastfood) \cite{le2013fastfood}: It uses the Hadamard transform to speed up the matrix multiplication in RFE. 
\item Binary Johnson-Lindenstrauss Embedding (BJLE) : It composes JL embedding with sign function (i.e., $\mathbf{z} = \text{sign}(\mathbf{Rx})$) which is a common binary embedding method.
\item Binary Codes for Shift-Invariant Kernels (BCSIK) \cite{raginsky2009locality}: It composes the Random Fourier Features from RFE with random sign function.
\item Our proposed method with full-precision model coefficients
\item Our proposed method with binary model coefficients 
\end{itemize}
To evaluate the performance of these six methods, we use five real-world benchmark datasets. The detailed information about these datasets is summarized in Table \ref{table.data}.  The first three datasets are download from LIBSVM website \footnote{\url{https://www.csie.ntu.edu.tw/~cjlin/libsvmtools/datasets/}}. {\sf mnist} and {\sf fashion-mnist} are download from openml \footnote{\url{https://www.openml.org/search?type=data}}

\begin{table}[h]
	\centering
	\caption{Experiment datasets}
	\begin{tabular}{ccccc}
		\hline
		Dataset & class &train size & test size & d     \\
		\hline
		{\sf usps}&10	  & 7,291	& 2007 &  256    \\
		{\sf covtype}   &2   & 464,810  & 114,202&  54\\
		{\sf webspam}   &2   &  280,000& 70,000 &  254\\
		{\sf mnist}   &10   & 60,000  & 10,000 &  780 \\
		{\sf fashion-mnist}  &10   & 60,000  & 10,000 &  780 \\
		\hline
		\label{table.data}
	\end{tabular}	
\end{table}

\subsection{Experiment results}
All the data are normalized by min-max normalization such that the feature values are within the range $[-1, 1]$. The dimension of nonlinear feature mapping $p$ is set to 2048. The $\sigma$ is chosen from $\{2^{-5}, 2^{-4}, \dots, 2^5\}$. The regularization parameter in both linear SVM and our method is chosen from $\{10^{-3}, 10^{-2}, \dots, 10^3\}$. The prediction accuracy and the memory cost for model inference for all algorithms are reported in Table \ref{tab:results}. Memory reduction is also included in Table \ref{tab:results}, where 86x means the memory cost is 86 times smaller compared with RFE. As shown in Table \ref{tab:results}, RFE gets the best classification accuracy for all five datasets. However, the memory cost for model inference is very high. The binary embedding methods BJLE and BCSIK can only reduce the memory cost by a small amount because the bottleneck is the transformation matrix $\mathbf{R} \in \mathbb{R}^{d\times p}$. Compared with BJLE and BCSIK, Fastfood is memory-efficient for nonlinear feature mapping. However, the memory cost for the classification model could be high for Fastfood, especially when the number of classes is large. Compared with RFE, our proposed method with ternary coefficients can significantly reduce the memory cost from 19x to 243x for model inference.

In next, we explore different properties of our proposed algorithm on {\sf usps} dataset with $\sigma$ set to $2^{-1}$. 

\begin{figure}[!h]
    \centering
    \subfigure[Memory]{\label{Fig.1.0}\includegraphics[width=0.45\columnwidth]{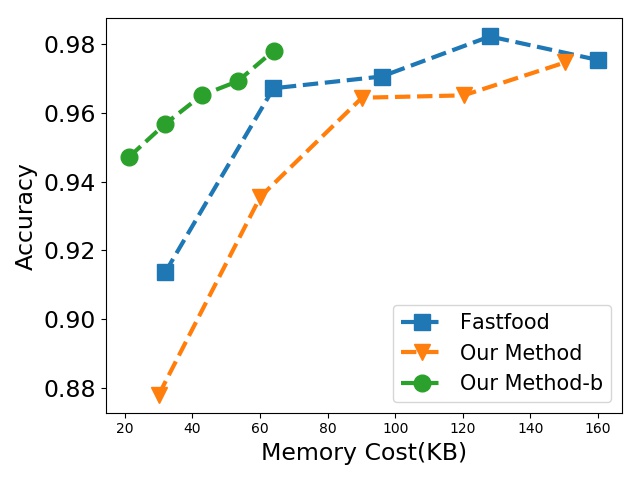}}
    \subfigure[Computation time]{\label{Fig.1.add}\includegraphics[width=0.45\columnwidth]{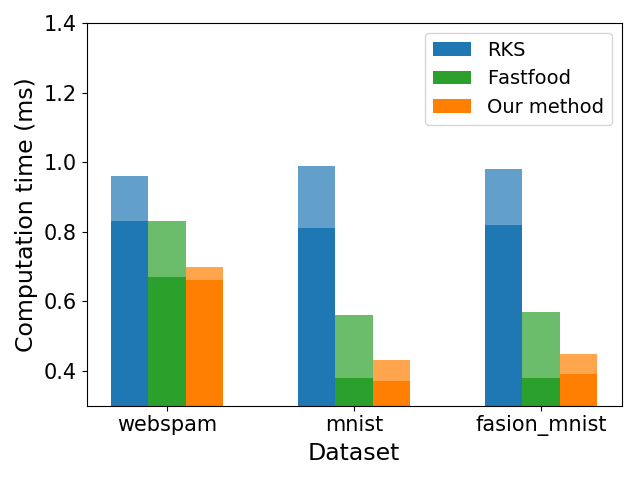}}
    \caption{Memory and computation efficiency}
    \label{fig:acc_meml}
\end{figure}

\textbf{Memory Efficiency of Binary Model}  Figure \ref{Fig.1.0} illustrates the impact of parameter $p$. As can be seen from it, the accuracy increases as $p$ increases and will converge if $p$ is large enough.  Besides, we further compare the memory cost and the prediction accuracy of three Fastfood-based kernel approximation methods.
 We take the usps dataset as an example. We can set larger p to gain the same performance of the full precision methods shown in Table.\ref{tab:results}. Furthermore, if we consider the memory cost, our binary model can achieve higher accuracy with the same memory, as shown in \ref{Fig.1.0}. 

\textbf{Computation Time}
We compare the time consumption of our method to process one single sample with the processing time of RFF and the original Fastfood kernel approximation method in Fig.\ref{Fig.1.add}.  The dark areas represent the feature embedding time, and the light areas are the prediction time. We show that our method's total computation time can be significantly reduced compared with the other two methods. We have two main observations. First, the Fast Fourier transform in random projection can significantly reduce the projection time, especially when the original data is a high-dimensional dataset (e.g., MNIST and Fashion MNIST). Besides, the binary embedding will add few additional time in feature embedding but will significantly improve the speed in the prediction stage. 

\begin{figure}[h]
    \centering
    \subfigure[Convergence] {\label{Fig.1.2}\includegraphics[width=0.45\columnwidth]{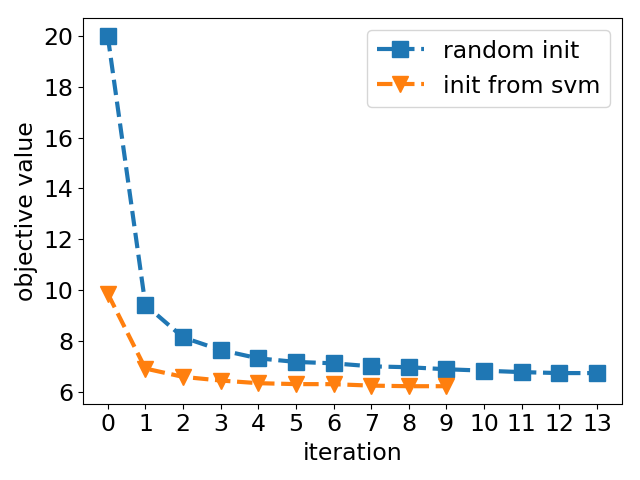}}
    \subfigure[Accuracy] {\label{Fig.1.3}\includegraphics[width=0.45\columnwidth]{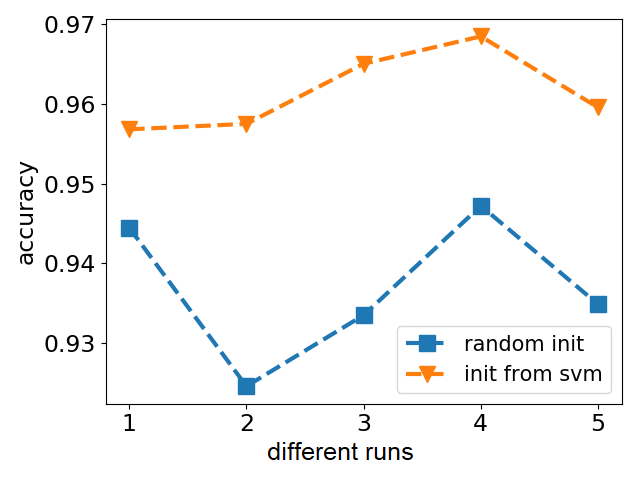}}
    \caption{Comparison between random initialization and our proposed initialization method}
    \label{fig:intialization}
\end{figure}

\textbf{Convergence of Our Algorithm.} In Fig.\ref{Fig.1.2}, we empirically show how our proposed algorithm converges. Here, we compare the two different initialization methods: (1)  random initialization; (2) initialization from linear SVM solution. We can observe that using the initialization from a linear SVM solution leads to a slightly lower objective value and converges in a few iterations. Motivated by this observation, we train a linear model on a small subset of data and binarize it as the initial $\mathbf{w}$ for our algorithm in practice. 

\textbf{Effectiveness of SVM initialization} In Fig.\ref{Fig.1.3}, we further illustrate the effect of our initialization strategy by compare the prediction accuracy. We can observe that using the initialization from a linear SVM solution leads to a higher accuracy and more stable compare with the random initialization.



\begin{figure}[!h]
    \centering
    \subfigure[binary coefficients]{\label{Fig.binary}\includegraphics[width = 0.45\columnwidth]{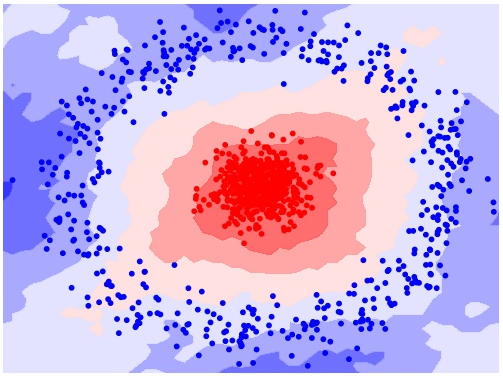}}
    \subfigure[ternary coefficients]{\label{Fig.binary}\includegraphics[width = 0.45\columnwidth]{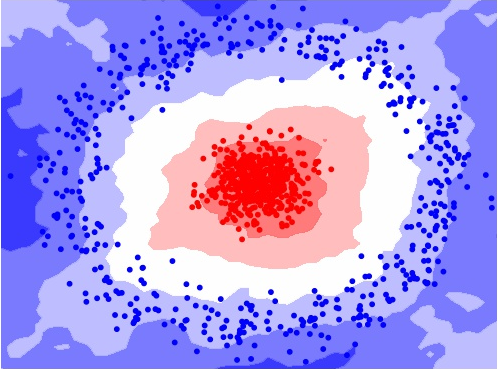}}
    \caption{Comparison of the decision boundaries between binary coefficients and ternary coefficients }
    \label{fig:dicision_boundary}
\end{figure}

\textbf{Decision Boundary of Ternary Coefficients}
We use the synthetic nonlinear \textit{circle} dataset to illustrate the effect of the ternary coefficients. The \textit{circle} is in two-dimensional space as shown in Figure \ref{fig:dicision_boundary}. The blue points in the larger outer circle belong to one class, and the red ones belong to another. We show the decision boundary of using binary and ternary coefficients. As shown in this figure, our proposed binary embedding with linear classifiers can produce effective nonlinear decision boundaries. Besides, as the feature binary embedding might involve some additional noise, the classification model using ternary coefficients can produce better and smoother decision boundary than using binary coefficients.

\section{Conclusion} 
This paper proposes a novel binary embedding method that can preserve the kernel similarity among data samples. Compared to BCSIK, our proposed method reduces the memory cost from $O(dp)$ to $O(p)$ and the computation cost from $O(dp)$ to $O(plog(d))$ for binary embedding. Besides, we propose a new algorithm to learn the classification model with ternary coefficients. Our algorithm can achieve better generalization accuracy than existing works on learning binary coefficients since we allow coefficient to be \{$-1, 0, 1$\} during the training stage. Our proposed algorithm can be easily implemented and applicable to other types of loss function and regularizer. We also provide a detailed analysis of the convergence of our algorithm and the inference complexity of our model. We evaluate our algorithm based on five large benchmark datasets and demonstrate our proposed model can build accurate nonlinear SVM models with memory cost less than 30KB on all five datasets. 

\bibliographystyle{aaai21}
\bibliography{BinarizedKSVM}

\end{document}